\documentclass[11pt,a4paper]{article}
\usepackage[hyperref]{acl2020}
\usepackage{times}
\usepackage{latexsym}
\usepackage{comment}

\usepackage{url}
\usepackage{amsmath}

\usepackage{CJKutf8}
\usepackage[boxed]{algorithm}
\usepackage{amssymb}

\usepackage{caption}
\usepackage{algpseudocode}

\aclfinalcopy 
\def\aclpaperid{59} 

\newenvironment{tightitemize}%
  {\begin{itemize}[topsep=0pt, partopsep=0pt] %
    \setlength{\itemsep}{0pt}%
    \setlength{\parskip}{0pt}%
    }%
  {\end{itemize}}
\usepackage{booktabs}
\usepackage{subcaption}
\usepackage{times}
\usepackage{url}
\usepackage{latexsym}
\usepackage{xspace}
\usepackage{amstext}
\usepackage{booktabs}
\usepackage{color}
\usepackage{graphicx}
\usepackage{tabulary}
\usepackage{graphicx}
\usepackage{enumitem}
\usepackage{lipsum}
\usepackage{amsfonts}

\usepackage{color}

\title{A Unified MRC Framework for Named Entity Recognition}

\author{Xiaoya Li$^{\clubsuit}$, Jingrong Feng$^{\clubsuit}$, Yuxian Meng$^{\clubsuit}$, Qinghong Han$^{\clubsuit}$ \\
{\bf Guoyin Wang$^{\blacklozenge}$, Fei Wu$^{\spadesuit}$ and Jiwei Li$^{\clubsuit}$}  \\
$^{\clubsuit}$ Shannon.AI, $^{\blacklozenge}$Amazon  ~~\\
$^{\spadesuit}$ Department of Computer Science and Technology, Zhejiang University\\
  \{xiaoya\_li, jingrong\_feng, yuxian\_meng,qinghong\_han\}@shannonai.com\\
   wufei@cs.zju.edu.cn,  jiwei\_li@shannonai.com
}

\date{}

\begin{document}
\maketitle

\begin{abstract}
The task of named entity recognition (NER)  is normally divided into nested NER and flat NER depending  on whether named entities are nested or not. Models are usually separately developed for the two tasks, since   sequence labeling models are only able to assign a single label to a particular token, which is unsuitable for nested NER where a token may be assigned several labels.

In this paper, we propose a unified framework  that is capable of handling both flat and nested NER tasks. Instead of treating the task of NER as a sequence labeling problem, we propose to formulate it as a machine reading comprehension (MRC) task. For example, extracting entities with the \textsc{per(Person)} label is formalized as extracting answer spans to the question  ``which person is mentioned in the text".This formulation naturally  tackles the entity overlapping issue in nested NER: the extraction of two overlapping entities with different categories requires answering two independent questions. Additionally, since the query encodes informative prior knowledge, this strategy facilitates the process of entity extraction, leading to better performances for not only nested NER, but flat NER.

We conduct experiments on both nested and flat NER datasets. Experiment results demonstrate  the effectiveness of the proposed formulation. We are able to achieve a vast amount of performance boost over current SOTA models on nested NER datasets, i.e.,   +1.28, +2.55, +5.44, +6.37,respectively on ACE04, ACE05, GENIA and KBP17, as well as flat NER datasets, i.e., +0.24, +1.95, +0.21, +1.49 respectively on English CoNLL 2003, English OntoNotes 5.0, Chinese MSRA and Chinese OntoNotes 4.0. The code and datasets can be found at \url{ https://github.com/ShannonAI/mrc-for-flat-nested-ner}.

\end{abstract}

\section{Introduction}

\begin{figure}
\includegraphics[width=3in]{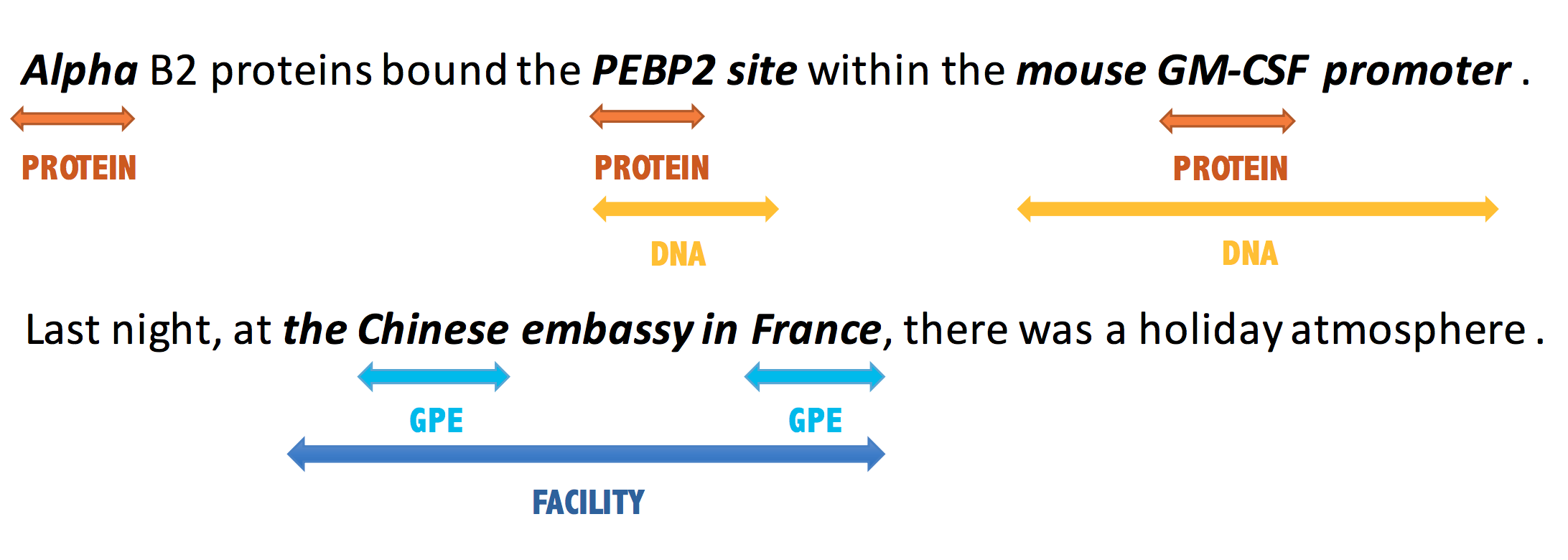}
\centering
\caption{Examples for {\em nested} entities from GENIA and ACE04 corpora. } 
\label{illustration}
\end{figure}

Named Entity Recognition (NER)  refers to the task of detecting the span and the semantic category of entities from a chunk of text.
The task can be further divided into two sub-categories, nested NER and flat NER,  
depending  on whether entities are nested or not. 
Nested NER refers to a phenomenon that the spans of entities (mentions) are nested, 
as shown in Figure~\ref{illustration}. Entity overlapping is a fairly common phenomenon in natural languages.   

The task of flat NER is commonly formalized as a sequence labeling task:
a 
sequence labeling model \cite{chiu2016named, ma2016end, devlin2018bert}
is trained to assign a single tagging class to each unit within a sequence of tokens. 
This formulation is unfortunately incapable of handling overlapping entities in nested NER \cite{huang2015bidirectional, chiu2015named}, where
multiple categories
 need to be assigned  to a single token if the token
 participates in multiple entities. 
Many attempts have been made to reconcile sequence labeling models with nested NER \cite{alex2007recognising,byrne2007nested,finkel2009nested,luroth2015,kati2018nested},  mostly based on the pipelined systems.
However, pipelined systems suffer from
the disadvantages of
error propagation, long running time and the
intensiveness 
in developing hand-crafted features, etc.

 Inspired by the current trend of formalizing NLP problems  as  
question answering tasks \cite{levy2017zero,mccann2018natural, xiaoya2019relation},
we propose a new framework that is capable of handling both flat and nested NER. 
Instead of treating the task of NER as a sequence labeling problem, we propose to formulate it as a SQuAD-style \cite{rajpurkar2016squad, rajpurkar2018know} machine reading comprehension (MRC) task.
 Each entity type is characterized by a natural language query, and entities are extracted by answering these queries given the contexts. 
For example, the task of assigning the \textsc{PER(person)} label to ``{\em [Washington] was born into slavery on the farm of James Burroughs\/}'' 
 is formalized as answering the question ``{\it which person is mentioned in the text?}''. 
This strategy naturally  tackles the entity 
overlapping issue in nested NER: the extraction of two  entities with different categories that overlap requires answering two independent questions. 

The MRC formulation also comes with another key  advantage over the sequence labeling formulation. 
For the latter,
golden NER categories are merely class indexes and lack for semantic prior information 
for
 entity categories. 
For example, 
the 
 \textsc{ORG(organization)} class is treated as 
  a one-hot vector in sequence labeling training. 
This lack of clarity on what to extract leads to inferior performances. 
On the contrary, for the MRC formulation,
the  query
encodes significant prior information about the entity category to extract. 
For example, the query  {\em ``find an organization such as company, agency and institution in the context"}   encourages the model to link
the word
 {\em ``organization"} in the query 
   to location entities in the context. 
   Additionally, by encoding comprehensive descriptions (e.g., 
{\em ``company, agency and institution"}) of tagging categories  (e.g., {\em ORG}), 
the model has the potential to 
    disambiguate  similar tagging classes.

We conduct experiments on both {\em nested} and {\em flat} NER datasets  to show the generality of our approach. 
Experimental results demonstrate its effectiveness. We are able to achieve a vast amount of performance boost over current SOTA models on nested NER datasets, i.e., 
+1.28, +2.55, +5.44, +6.37,
 respectively on ACE04, ACE05, GENIA and KBP17, 
 as well as flat NER datasets, i.e.,
 +0.24, +1.95, +0.21, +1.49 respectively on English CoNLL 2003, English OntoNotes 5.0, Chinese MSRA, Chinese OntoNotes 4.0. 
We wish that our work would inspire the introduction of new  paradigms for the entity recognition task.

\section{Related Work}
\label{related section}
\subsection{Named Entity Recognition (NER)} 
Traditional  sequence  labeling  models
use CRFs \cite{lafferty2001conditional,sutton2007dynamic} as a backbone for NER.
The first work using neural models for NER goes back to 2003, when
\newcite{hammerton2003named} attempted to solve the problem using unidirectional LSTMs. 
\newcite{collobert2011natural} presented a CNN-CRF structure, augmented  with character embeddings by \newcite{santos2015boosting}. 
\newcite{lample2016neural}  explored neural structures for NER, in which
the bidirectional LSTMs are combined with CRFs  
 with features based on 
 character-based word
representations  and unsupervised word representations.
\newcite{ma2016end} 
 and \newcite{chiu2016named}  
 used a character CNN to extract features from characters.
  Recent large-scale language model pretraining methods such as BERT \cite{devlin2018bert} and ELMo \cite{elmo2018} further enhanced the performance of NER, yielding  state-of-the-art performances. 

\subsection{Nested Named Entity Recognition} 

The overlapping between entities (mentions)  was first noticed by \newcite{kim2003genia}, who 
developed handcrafted 
 rules  to identify overlapping mentions. 
 \newcite{alex2007recognising} proposed two multi-layer CRF models for nested NER. The first model is the inside-out model, in which the first CRF  identifies the innermost
entities, and the successive layer CRF is built over words
and the innermost entities extracted from the previous CRF
to identify second-level entities, etc. The other is  
the outside-in model, in which the first CRF
 identifies outermost entities, and then successive CRFs would identify increasingly nested entities.
\newcite{finkel2009nested} built a model to extract nested entity mentions based on parse trees. 
They made the assumption that one mention is fully contained by the other when they overlap. 
\newcite{luroth2015} proposed to use mention hyper-graphs for recognizing overlapping mentions. 
\newcite{xu2017fofe} utilized a local classifier that runs on every possible span to detect overlapping mentions and 
\newcite{kati2018nested}
used neural models to learn the  hyper-graph representations for nested entities. 
\newcite{ju-etal-2018-neural} dynamically stacked flat NER layers in a hierarchical manner.
\newcite{nugget2019nested} proposed the Anchor-Region Networks (ARNs) architecture by modeling and leveraging the head-driven phrase structures of nested entity mentions.
\newcite{luan2019dygie} built a span enumeration approach by selecting the most confident entity spans and linking these nodes with confidence-weighted relation types and coreferences. 
Other works \cite{muis-lu-2017-labeling,sohrab-miwa-2018-deep,zheng-etal-2019-boundary} also proposed various methods to tackle the nested NER problem.

Recently, nested NER models
are enriched with
 pre-trained contextual embeddings such as BERT \cite{devlin2018bert} and ELMo \cite{peters2018deep}.
\newcite{joseph2019merge} introduced a BERT-based model  that first merges tokens and/or entities into entities, and then assigned labeled to these entities. 
\newcite{shibuya2019second}  provided inference model that extracts entities
iteratively from outermost ones to inner ones. 
\newcite{strakova2019nested} viewed nested NER as a sequence-to-sequence generation  problem, in which the input sequence is a list of tokens and the target sequence is a list of labels.

\subsection{Machine Reading Comprehension (MRC)}
MRC models  \cite{seo2016bidirectional,wang2016multi,wang2016machine,xiong2016dynamic,xiong2017dcn,wang2016multi,shen2017reasonet,chen2017reading} extract answer spans from a passage through a given question. 
The task can be formalized as two multi-class classification tasks, i.e., predicting the starting and ending positions of the answer spans.

Over the past one or two years, 
there has been a trend of 
 transforming NLP tasks to MRC question answering. For example, \citet{levy2017zero} 
transformed the task of relation extraction to a QA task:
each relation type $R(x,y)$ can be parameterized as a question $q(x)$ whose answer is $y$. For example, the relation \textsc{educated-at}
can be mapped to  “{\it Where did x study?}”. 
Given a question $q(x)$, 
if a non-null answer $y$ can be extracted from a sentence, it means 
the relation label for the current sentence is $R$. 
\newcite{mccann2018natural} transformed 
 NLP tasks such as summarization or sentiment analysis 
    into question answering. 
For example, the task of summarization can  be formalized as answering the question ``{\it What is the
summary?}''. 
Our work is significantly inspired by \newcite{xiaoya2019relation}, which formalized the task of entity-relation extraction as a multi-turn question answering task.
Different from this work, \newcite{xiaoya2019relation} focused on relation extraction rather than NER.
Additionally,  
\newcite{xiaoya2019relation} utilized a template-based procedure for constructing queries to extract semantic relations between entities and their queries lack  diversity. 
In this paper, more factual knowledge such as synonyms and examples are incorporated into queries, and
we  present an in-depth analysis of the impact of strategies of building queries. 

\section{NER as MRC}
\subsection{Task Formalization}
Given an input sequence $X = \{x_1, x_2, ..., x_n\}$, where $n$ denotes the length of the sequence, we need to find every entity in $X$, and then assign a label $y \in Y$ to it, where $Y$ is a predefined list of all possible tag types (e.g., PER, LOC, etc). 

\paragraph{Dataset Construction}
Firstly we  need to transform the tagging-style annotated NER dataset  to a set of 
\textsc{(question, answer, context)} triples. 
For each tag type $y\in Y$, it is associated with a natural language question $q_y = \{q_1, q_2, ..., q_m\}$, where $m$ denotes the length of the generated query. 
 An annotated entity $x_\text{start,end}=\{x_\text{start}, x_\text{start+1},\cdots,x_\text{end-1},x_\text{end}\}$ is a substring of $X$ satisfying $\text{start}\le\text{end}$. Each entity is associated 
 with a golden label $y\in Y$. By generating a natural language question $q_y$ based on the label $y$, 
we can obtain  the triple ($q_y$, $x_\text{start,end}$, $X$), which is exactly the \textsc{(question, answer, context)} triple that we need. Note that we use the subscript ``$_\text{start,end}$'' to denote the continuous tokens from index `start' to  `end' in a sequence. 

\subsection{Query Generation}
The question generation procedure is important since queries encode prior knowledge about labels and have a significant influence on the final results. 
Different ways have been proposed for question generation, e.g., 
\newcite{xiaoya2019relation} utilized a template-based procedure for constructing queries to extract semantic relations between entities. 
In this paper, we take annotation guideline notes as references to construct queries. 
Annotation guideline notes are the guidelines provided to the annotators of the dataset 
by the dataset builder. They are  descriptions of tag categories, which are described as generic and precise as possible so that human annotators can annotate the concepts or mentions in any text without running into ambiguity. Examples are shown in Table \ref{entitytemplate}.

\subsection{Model Details} 
\subsubsection{Model Backbone}
Given the question $q_y$,
we need to extract the text span $x_\text{start,end}$ which is with type $y$  from $X$ under the MRC framework.  
We use BERT \cite{devlin2018bert} as the backbone. 
To be in line with BERT, the question  $q_y$ and the passage $X$ are concatenated, forming the combined string 
 $\{[\text{CLS}], q_1, q_2, ..., q_m, [\text{SEP}], x_1, x_2, ..., x_n\}$, where [CLS] and [SEP] are special tokens. Then BERT receives the combined string and outputs a context representation matrix $E\in\mathbb{R}^{n\times d}$, where $d$ is the vector dimension of the last layer of BERT and we simply drop the query representations. 
 
\subsubsection{Span Selection} There are  two strategies  for span selection in MRC:
the first strategy \cite{seo2016bidirectional,wang2016multi} is to have two $n$-class classifiers separately predict the  start index and the end index, where $n$ denotes the length of the context.
Since the softmax function is put over all tokens in the context, this strategy has the disadvantage of only being able to output a single span given a query; 
the other strategy is to have two binary classifiers, 
one to predict whether each  token is the start index or not, 
the other to predict whether each token is the end index or not. 
This strategy allows for outputting multiple start indexes and multiple end indexes for a given context and a specific query, and thus has the potentials to extract all related entities according to $q_y$.
We adopt the second strategy and describe the details below.

\begin{table}[t]
\center
\small
\begin{tabular}{ll}\hline
{\bf Entity} & {\bf Natural Language Question} \\\hline
Location  & Find locations in the text,  including  non-\\
&geographical locations, mountain ranges
\\
&and bodies of water.  \\
Facility & Find facilities in the text, including \\
&buildings,  airports, highways and bridges.  \\ 
Organization  & Find organizations  in the text, including \\
& companies, agencies and institutions.  \\\hline
\end{tabular}
\caption{Examples for transforming different entity categories to question queries. }
\label{entitytemplate}
\end{table}

\paragraph{Start Index Prediction}
Given the 
representation matrix 
$E$ output from BERT, the model first  predicts the probability of each token being a start index as follows:
\begin{equation}
    P_\text{start}=\text{softmax}_\text{each row}(E\cdot T_\text{start})\in\mathbb{R}^{n\times 2}
\end{equation}
$T_\text{start}\in\mathbb{R}^{d\times 2}$ 
 is the weights to learn. Each row of $P_\text{start}$ presents the probability distribution of each index being the start position of an entity given the query. 
 
 \paragraph{End Index Prediction}
The end index prediction procedure is exactly the same, except that we have another matrix $T_\text{end}$ to obtain probability matrix $P_\text{end}\in\mathbb{R}^{n\times 2}$.

\paragraph{Start-End Matching}
In the  context $X$, there could be multiple entities of the same category. This means
that 
 multiple start indexes could be predicted from the {\it start-index prediction} model and multiple end indexes predicted from the {\it end-index prediction} model. The heuristic of matching the start index with its nearest end index does not work here since entities could overlap. 
We thus further need a method to match a predicted start index with its corresponding end index. 

Specifically, by applying argmax to each row of $P_\text{start}$ and $P_\text{end}$, we will get
the predicted indexes that might be the starting or ending positions, i.e., 
 $\hat{I}_\text{start}$  and $\hat{I}_\text{end}$:
\begin{equation}
\begin{aligned}
\small
&\hat{I}_\text{start}=\{i~|~\text{argmax}(P_\text{start}^{(i)})=1,i=1,\cdots,n\}\\
&\hat{I}_\text{end}=\{j~|~\text{argmax}(P_\text{end}^{(j)})=1,j=1,\cdots,n\}
\end{aligned}
\end{equation}
where the superscript $^{(i)}$ denotes the $i$-th row of a matrix. Given any start index $i_\text{start}\in \hat{I}_\text{start}$ and end index  $i_\text{end}\in \hat{I}_\text{end}$, 
a binary classification model is trained to predict 
the probability that they should be matched, given as follows: 
\begin{equation}
\label{equ4}
    P_{i_\text{start},j_\text{end}}=\text{sigmoid}(m\cdot\text{concat}(E_{i_\text{start}}, E_{j_\text{end}}))
\end{equation}
where $m\in\mathbb{R}^{1\times 2d}$ is the weights to learn.

\subsection{Train and Test}
At training time, $X$ is  paired with two 
label sequences $Y_\text{start}$ and $Y_\text{end}$ of length $n$ 
representing the ground-truth label of each token $x_i$ being the start index or end index of any entity. We therefore have the following two losses for start and end index predictions:
\begin{equation}
\begin{aligned}
    &\mathcal{L}_\text{start}=\text{CE}(P_\text{start}, Y_\text{start}) \\
    &\mathcal{L}_\text{end}=\text{CE}(P_\text{end}, Y_\text{end})
 \end{aligned}
\end{equation}
Let $Y_\text{start, end}$  denote the golden labels for whether each start index should be matched with each end index. The start-end index matching loss is given as follows:
\begin{equation}
    \mathcal{L}_\text{span}=\text{CE}(P_{\text{start},\text{end}}, Y_\text{start, end})
    \label{training-obj}
\end{equation}
The overall training objective to be minimized is as follows:
\begin{equation}
    \mathcal{L}=\alpha\mathcal{L}_\text{start}+\beta\mathcal{L}_\text{end}+\gamma\mathcal{L}_\text{span}
\end{equation}
$\alpha,\beta,\gamma\in [0,1]$ are hyper-parameters to control the contributions towards the overall training objective. The three losses are jointly trained in an end-to-end fashion, with parameters shared at the BERT layer. At test time, start and end indexes are first separately selected based on $\hat{I}_\text{start}$ and $\hat{I}_\text{end}$. Then the index matching model is used to align the extracted start indexes with end indexes, leading to the final extracted answers. 


\section{Experiments}

\subsection{Experiments on Nested NER} 
\subsubsection{Datasets} 
For {\em nested} NER, experiments are conducted on the  
widely-used  ACE 2004, ACE 2005, GENIA and KBP2017 datasets, which respectively contain  24\%, 22\%, 10\% and 19\% nested mentions. 
Hyperparameters are tuned on their corresponding development sets. 
For evaluation, we 
use span-level micro-averaged precision, recall and F1.  

\paragraph{ACE 2004 and ACE 2005}\cite{ace2004ner, ace2005ner}:
The two datasets each contain 7 entity categories. 
For each entity type, there are annotations for
both
 the entity mentions and mention heads. 
For fair comparison, 
we exactly follow the data preprocessing strategy in \newcite{kati2018nested} and \newcite{hongyu2019nugget} by keeping files from \texttt{bn}, \texttt{nw} and \texttt{wl},  and splitting these files into train, dev and test sets  by 8:1:1, respectively.

\paragraph{GENIA} \cite{genia} For the GENIA dataset, we use GENIAcorpus3.02p.
We follow the protocols in \newcite{kati2018nested}.

\paragraph{KBP2017} We follow \newcite{kati2018nested} and evaluate our model on the 2017 English evaluation dataset (LDC2017D55). 
Training set consists of RichERE annotated datasets, which include LDC2015E29, LDC2015E68, LDC2016E31 and LDC2017E02. 
We follow 
 the dataset split strategy in  \newcite{hongyu2019nugget}.

\subsubsection{Baselines} 
We use the following models as baselines:
\begin{tightitemize}
\item {\bf Hyper-Graph:} \newcite{kati2018nested} proposes a hypergraph-based model based on LSTMs. 
\item {\bf Seg-Graph:} \newcite{luwei2018overlap} proposes a  segmental hypergargh representation to model overlapping entity mentions. 
\item {\bf ARN:} \newcite{nugget2019nested} proposes Anchor-Region Networks by modeling and levraging the head-driven phrase structures of entity mentions. 
\item {\bf KBP17-Best:} \newcite{heng2017kbp} gives an overview of the Entity Discovery task at the Knowledge Base Population (KBP) track at TAC2017 and also reports previous best results for the task of nested NER. 
\item {\bf Seq2Seq-BERT:} \newcite{strakova2019nested} views the {\em nested} NER as a sequence-to-sequence problem. Input to the model is word tokens and the output sequence consists of labels.  
\item {\bf Path-BERT:} \newcite{shibuya2019second} treats the tag sequence as the second best path within in the span of their parent entity based on BERT. 
\item {\bf Merge-BERT:} \newcite{joseph2019merge} proposes a merge and label method based on BERT. 
\item {\bf DYGIE:} \newcite{luan2019dygie} introduces a general framework that share span representations using dynamically constructed span graphs. 
\end{tightitemize}
 \begin{table}
\resizebox{\columnwidth}{58mm}{
\begin{tabular}{llll}
\toprule
\multicolumn{4}{c}{{\bf English ACE 2004}}\\
\midrule
{\bf Model} & {\bf Precision} & {\bf Rrecall} & {\bf F1} \\
\midrule
Hyper-Graph \cite{kati2018nested} & 73.6 & 71.8 & 72.7 \\
Seg-Graph \cite{luwei2018overlap} & 78.0 & 72.4 & 75.1 \\
Seq2seq-BERT \cite{strakova2019nested} & - &- &  84.40\\ 
Path-BERT \cite{shibuya2019second} & 83.73 & 81.91 & 82.81 \\ 
DYGIE \cite{luan2019dygie} & - & - & 84.7 \\
BERT-MRC& {\bf 85.05} & {\bf 86.32} & {\bf 85.98} \\
& & & {\bf (+1.28)}\\
\bottomrule 
\multicolumn{4}{c}{{\bf English ACE 2005}}\\
\midrule 
{\bf Model} & {\bf Precision} & {\bf Recall} & {\bf F1} \\
\midrule 
Hyper-Graph \cite{kati2018nested} & 70.6 & 70.4 & 70.5 \\
Seg-Graph \cite{luwei2018overlap} & 76.8 & 72.3 & 74.5 \\
ARN \cite{nugget2019nested} & 76.2 & 73.6 & 74.9 \\
Path-BERT \cite{shibuya2019second} & 82.98 & 82.42 & 82.70 \\ 
Merge-BERT \cite{joseph2019merge} & 82.7 & 82.1 & 82.4 \\
DYGIE \cite{luan2019dygie} & - & - & 82.9 \\
Seq2seq-BERT \cite{strakova2019nested} & - &- &  84.33 \\ 
BERT-MRC& {\bf 87.16} & {\bf 86.59} & {\bf 86.88} \\
&  & & {\bf (+2.55)}\\
\bottomrule 
\multicolumn{4}{c}{{\bf English GENIA}}\\
\midrule 
{\bf Model} & {\bf Precision} & {\bf Recall} & {\bf F1} \\
\midrule 
Hyper-Graph \cite{kati2018nested} & 77.7 & 71.8 & 74.6 \\
ARN \cite{nugget2019nested} & 75.8 & 73.9 & 74.8 \\
Path-BERT \cite{shibuya2019second} & 78.07 & 76.45 & 77.25 \\ 
DYGIE \cite{luan2019dygie} & - & - & 76.2 \\
Seq2seq-BERT \cite{strakova2019nested} & - &- &  78.31 \\ 
BERT-MRC& {\bf 85.18} & {\bf 81.12} & {\bf 83.75} \\
&  & & {\bf (+5.44)}\\
\bottomrule 
\multicolumn{4}{c}{{\bf English KBP 2017}}\\
\midrule 
{\bf Model} & {\bf Precision} & {\bf Recall} & {\bf F1} \\
\midrule 
KBP17-Best \cite{heng2017kbp} & 76.2 & 73.0 & 72.8 \\
ARN \cite{nugget2019nested} & 77.7 & 71.8 & 74.6 \\
BERT-MRC& {\bf 82.33} & {\bf 77.61} & {\bf 80.97} \\
&  &  & {\bf (+6.37)}\\
\bottomrule \hline
\end{tabular}
}
\caption{Results for {\em nested} NER tasks.}
\label{nested NER results}
\end{table}

\subsubsection{Results} 
Table \ref{nested NER results} shows  experimental results on {\em nested} NER datasets. 
We observe huge performance boosts on the nested NER datasets over previous state-of-the-art models, achieving F1 scores of 
85.98\%, 86.88\%, 83.75\% and 80.97\% on ACE04, ACE05, GENIA and KBP-2017 datasets, which are 
+1.28\%, +2.55\%, +5.44\% and +6.37\% over previous SOTA performances, respectively. 

\subsection{Experiments on Flat NER}


\subsubsection{Datasets}
For {\em flat} NER, experiments are conducted on both English datasets i.e. CoNLL2003 and OntoNotes 5.0 and Chinese datasets i.e. OntoNotes 4.0 and MSRA. 
Hyperparameters are tuned on their corresponding development sets.
We  report span-level micro-averaged precision, recall and F1 scores for evaluation. 

\paragraph{CoNLL2003} \cite{conll2003ner} is an English dataset with four types of named entities: Location, Organization, Person and Miscellaneous. 
We followed data processing protocols in \newcite{ma2016end}.

\paragraph{OntoNotes 5.0} \cite{ontonotes5} is an English dataset and consists of text from a wide variety of sources.
The dataset includes 18 types of named entity, consisting of 11 types (Person, Organization, etc) and 7 values (Date, Percent, etc).  

\paragraph{MSRA} \cite{msra2006ner} is a Chinese dataset and performs as a benchmark dataset. 
Data in MSRA is collected from news domain and is used as shared task on SIGNAN backoff 2006. There are three types of named entities. 

\paragraph{OntoNotes 4.0} \cite{ontonotes4} is a Chinese dataset and consists of text from news domain. 
OntoNotes 4.0 annotates 18 named entity types. In this paper, we take the same data split as \newcite{wu2019glyce}. 

\subsubsection{Baselines}  

For English datasets, we use the following models as baselines. 
\begin{tightitemize}
\item {\bf BiLSTM-CRF} from \newcite{ma2016end}. 
\item {\bf ELMo} tagging model from \newcite{peters2018deep}. 
\item {\bf CVT} from \newcite{kevin2018cross}, which uses Cross-View Training(CVT) to improve the representations of a Bi-LSTM encoder. 
\item {\bf Bert-Tagger} from \newcite{devlin2018bert}, which treats NER as a tagging task.  
\end{tightitemize}
For Chinese datasets, we use the following models as baselines:
\begin{tightitemize}
\item {\bf Lattice-LSTM:} \newcite{latticelstm2018} constructs a word-character lattice. 
\item {\bf Bert-Tagger:} \newcite{devlin2018bert} treats NER as a tagging task. 
\item {\bf Glyce-BERT:} 
The current SOTA model in Chinese NER developed by \newcite{wu2019glyce}, which
 combines glyph information with BERT pretraining. 
\end{tightitemize}

\begin{table}[t]
\small
\centering
\resizebox{\columnwidth}{50mm}{
\begin{tabular}{llll}
\toprule
\multicolumn{4}{c}{{\bf English CoNLL 2003}}\\
\midrule
{\bf Model} & {\bf Precision} & {\bf Recall} & {\bf F1} \\
\midrule
BiLSTM-CRF \cite{ma2016end} & - & - & 91.03\\ 
ELMo \cite{peters2018deep} & - & - & 92.22 \\
CVT \cite{kevin2018cross} & -& - & 92.6 \\
BERT-Tagger \cite{devlin2018bert} & - & - & 92.8 \\
BERT-MRC& {\bf 92.33} & {\bf 94.61} & {\bf 93.04} \\
&  &  & {\bf (+0.24)}\\
\bottomrule
\multicolumn{4}{c}{{\bf English OntoNotes 5.0}}\\\midrule
{\bf Model} & {\bf Precision} & {\bf Recall} & {\bf F1} \\\midrule
BiLSTM-CRF \cite{ma2016end} & 86.04 & 86.53 & 86.28 \\
\newcite{dilatedcnn} & - & - & 86.84 \\
CVT \cite{kevin2018cross} & -& - & 88.8 \\
BERT-Tagger \cite{devlin2018bert} & 90.01& 88.35& 89.16 \\
BERT-MRC& {\bf 92.98} & {\bf 89.95} & {\bf 91.11} \\
&  &  & {\bf (+1.95)}\\
\bottomrule
\multicolumn{4}{c}{{\bf Chinese MSRA}}\\\midrule
{\bf Model} & {\bf Precision} & {\bf Recall} & {\bf F1} \\\midrule
Lattice-LSTM \cite{latticelstm2018}  & 93.57& 92.79 & 93.18  \\
BERT-Tagger \cite{devlin2018bert} & 94.97 & 94.62 & 94.80 \\
Glyce-BERT \cite{wu2019glyce}& 95.57 & 95.51 & 95.54\\
BERT-MRC& {\bf 96.18} & {\bf 95.12} & {\bf 95.75} \\
&  & & {\bf (+0.21)}\\
\bottomrule 
\multicolumn{4}{c}{{\bf Chinese OntoNotes 4.0}}\\\midrule
{\bf Model} & {\bf Precision} & {\bf Recall} & {\bf F1}  \\\midrule
Lattice-LSTM \cite{latticelstm2018}  & 76.35& 71.56& 73.88  \\
BERT-Tagger \cite{devlin2018bert} & 78.01& 80.35& 79.16 \\
Glyce-BERT \cite{wu2019glyce}& 81.87 & 81.40 & 81.63 \\
BERT-MRC& {\bf 82.98} & {\bf 81.25} & {\bf 82.11} \\
&  & & {\bf (+0.48)}\\\bottomrule 
\end{tabular}
}
\caption{Results for {\em flat} NER tasks.}
\label{flat NER results}
\end{table}

\subsubsection{Results and Discussions}Table \ref{flat NER results} presents  comparisons between the proposed model and  baseline models. For English CoNLL 2003, our model outperforms the fine-tuned BERT tagging model by +0.24\% 
in terms of F1, while for 
English OntoNotes 5.0, the proposed model achieves a huge gain of +1.95\% improvement. 
The reason why greater performance boost is observed for OntoNotes is that 
OntoNotes contains more types of entities than CoNLL03 (18 vs 4), and some entity categories face the severe data sparsity problem. 
Since the query  encodes significant prior knowledge for the entity type to extract, the MRC formulation is more immune to the tag sparsity issue, 
leading to more improvements on OntoNotes. 
The proposed method also achieves new state-of-the-art results on Chinese datasets. 
For Chinese MSRA, the proposed method outperforms the fine-tuned BERT tagging model by +0.95\% in terms of F1. 
We also improve the F1 from 79.16\% to 82.11\% on Chinese OntoNotes4.0.

\begin{figure*}[t]
\centering
\includegraphics[width=5.0in]{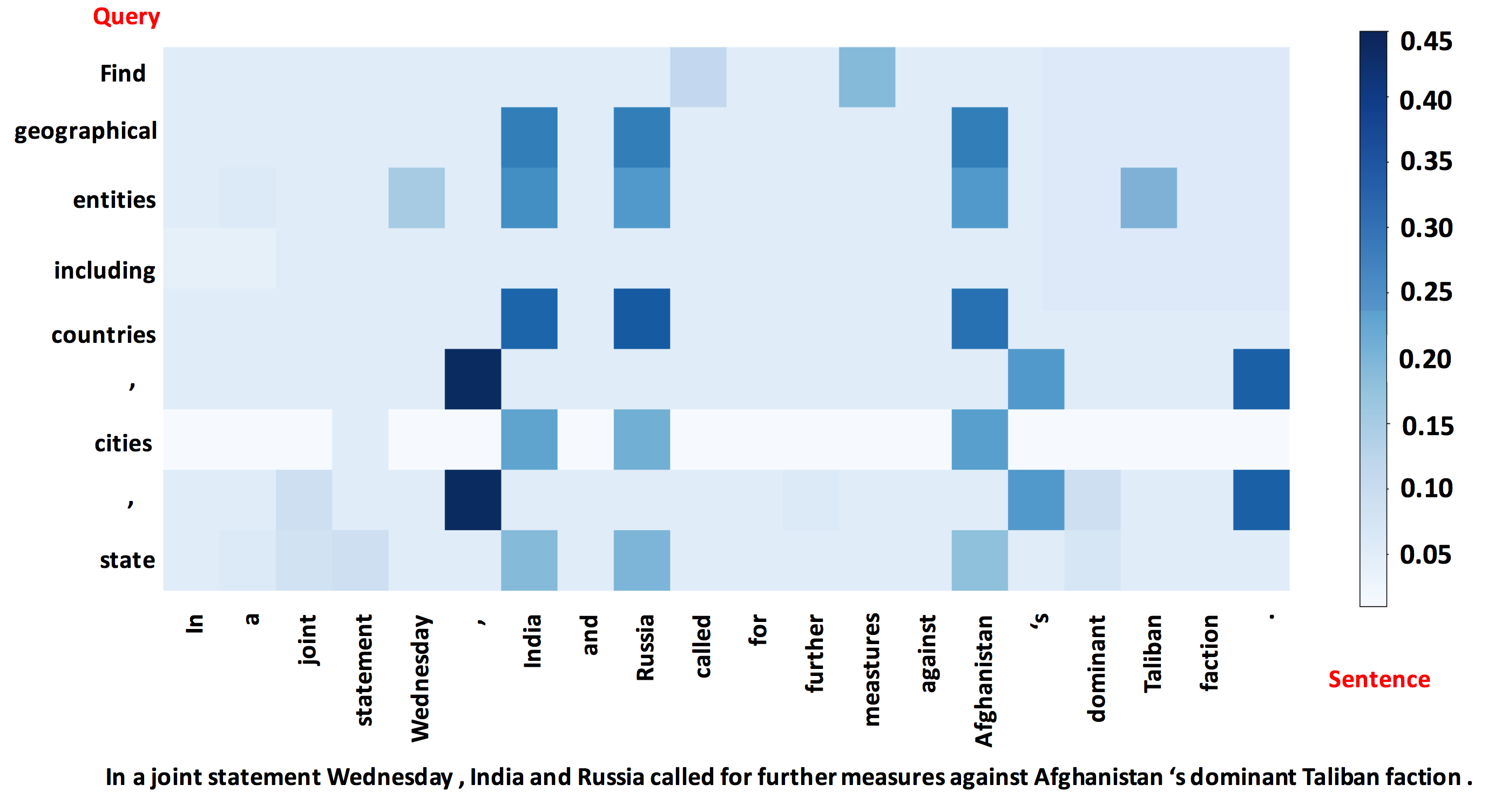}
\caption{An example of attention matrices between the query and the input sentence.} 
\label{headmap_abalation}
\end{figure*}

\section{Ablation studies}

\subsection{Improvement from MRC or from BERT} 
For flat NER, it is not immediately clear which proportion is responsible for the improvement, the MRC formulation or BERT \cite{devlin2018bert}. 
On one hand, the MRC formulation facilitates the entity extraction process by encoding prior knowledge in the query; on the other hand, the good performance might also come from the large-scale pre-training in BERT.

\begin{table}
\small
\center
\begin{tabular}{ll}\toprule
\multicolumn{2}{c}{{\bf English OntoNotes 5.0}}\\\midrule
{\bf Model} & {\bf F1} \\\midrule
LSTM tagger \cite{dilatedcnn} & 86.84 \\
BiDAF \cite{bidaf}  & 87.39 (+0.55) \\ 
QAnet \cite{qanet} &  87.98  (+1.14) \\ \hline
BERT-Tagger & 89.16 \\
BERT-MRC & {\bf 91.11} (+1.95) \\\bottomrule
\end{tabular}
\caption{Results of different MRC models on English OntoNotes5.0.}
\label{ablation-query-1}
\end{table}

To separate the influence from large-scale BERT pretraining, we compare the LSTM-CRF tagging model \cite{dilatedcnn} with other MRC based models such as QAnet \cite{qanet}
and BiDAF \cite{bidaf}, which do not rely on large-scale pretraining. 
Results on English Ontonotes   are shown in Table \ref{ablation-query-1}. 
As can be seen, though underperforming BERT-Tagger, 
the MRC based approaches 
QAnet and BiDAF still significantly outperform tagging models based on LSTM+CRF. This validates the 
importance of MRC formulation.
The MRC formulation's benefits are also verified when comparing BERT-tagger with  BERT-MRC: the latter outperforms the former by +1.95\%. 

We plot 
the attention matrices output from the BiDAF model between the query and the context sentence in 
 Figure \ref{headmap_abalation}. 
 As can be seen, the semantic similarity between tagging classes and the contexts are able to be captured in the  attention matrix. In the examples, {\em Flevland} matches {\em geographical}, {\em cities} and {\em state}.


\subsection{How to Construct Queries} 
\label{query strategy}
How to construct query has a significant influence on the final results. 
In this subsection, we explore different ways to construct queries and their influence, including:

\begin{table}
\small
\center
\begin{tabular}{ll}\toprule
\multicolumn{2}{c}{{\bf English OntoNotes 5.0}}\\\midrule
{\bf Model} &  {\bf F1} \\\midrule
BERT-Tagger & 89.16 \\\hline
{Position index of labels}  & 88.29 (-0.87) \\
{Keywords} &  89.74 (+0.58) \\
{Wikipedia} & 89.66 (+0.59)\\
{Rule-based template filling} & 89.30 (+0.14) \\
{Synonyms} &  89.92 (+0.76)\\
{Keywords+Synonyms}  &  90.23 (+1.07)\\
{Annotation guideline notes} & 91.11 (+1.95) \\\bottomrule
\end{tabular}
\caption{Results of different types of queries.}
\label{ablation-query-1}
\vskip -0.15in
\end{table}

\begin{tightitemize}
\item {\bf Position index of labels:} a query is constructed 
using the index of a tag to  
, i.e., "one", "two", "three".
\item {\bf Keyword:} a query is the keyword describing the tag, e.g., 
the question query for tag ORG is 
{\em ``organization"}.   
\item {\bf Rule-based template filling:} 
 generates 
  questions using templates. The query 
for tag ORG is
{\em ``which organization is mentioned in the text"}.   
\item {\bf Wikipedia:} a query is constructed using its  wikipedia definition. The query 
for tag ORG is
{\em "an organization is an entity comprising multiple people, such as an institution or an association."}

\item {\bf Synonyms:} are words or phrases that mean exactly or nearly the same as the original keyword extracted using
the Oxford Dictionary. The query 
for tag ORG is
{\em ``association"}.   

\item {\bf Keyword+Synonyms:} the concatenation of a keyword and its synonym.

\item {\bf Annotation guideline notes:} is the method we use in this paper. The query 
for tag ORG is {\em "find organizations including companies, agencies and institutions"}.

\end{tightitemize}

Table \ref{ablation-query-1} shows the experimental results on English OntoNotes 5.0. 
The BERT-MRC outperforms BERT-Tagger in all settings except  {\em Position Index of Labels}.
The model trained with the {\em Annotation Guideline Notes} achieves the highest F1 score.
Explanations are as follows: 
for {\em  Position Index Dataset}, 
queries are constructed using tag indexes and  thus do not contain any meaningful information, leading to inferior performances;  
{\em Wikipedia}  underperforms {\em Annotation Guideline Notes} because 
definitions from Wikipedia are relatively general and may not precisely describe the categories in a way tailored to data annotations.  



\begin{table}
\small
\center
\begin{tabular}{l|l|l|l}\toprule
{\bf Models} & {\bf Train} & {\bf Test} & {\bf F1} \\\bottomrule 
BERT-tagger &  OntoNotes5.0 & OntoNotes5.0  &89.16  \\
BERT-MRC &  OntoNotes5.0 & OntoNotes5.0  & 91.11  \\\hline
BERT-tagger &  CoNLL03 & OntoNotes5.0  & 31.87  \\
BERT-MRC &  CoNLL03 & OntoNotes5.0  &  72.34\\\bottomrule
\end{tabular}
\caption{Zero-shot evaluation on OntoNotes5.0. BERT-MRC can achieve better zero-shot performances.}
\label{zero-shot-table}
\end{table}

\subsection{Zero-shot Evaluation on Unseen Labels}
It would be interesting to test how well a model trained on one dataset is transferable to another, 
which is 
referred to as the zero-shot learning ability. 
We trained models on CoNLL 2003  
and test them on  OntoNotes5.0. 
OntoNotes5.0 contains 18 entity types, 3 shared with CoNLL03, and 15 unseen in CoNLL03. 
Table \ref{zero-shot-table} presents the results. As can been seen, BERT-tagger does not have zero-shot learning ability,
only obtaining an accuracy of 31.87\%.
This is in line with our expectation since it cannot 
predict  labels unseen from the training set. 
The question-answering 
formalization in 
MRC framework, which predicts the answer to the given query, comes with more generalization capability and achieves acceptable results. 

\subsection{Size of Training Data}

\begin{figure}
\includegraphics[width=3in]{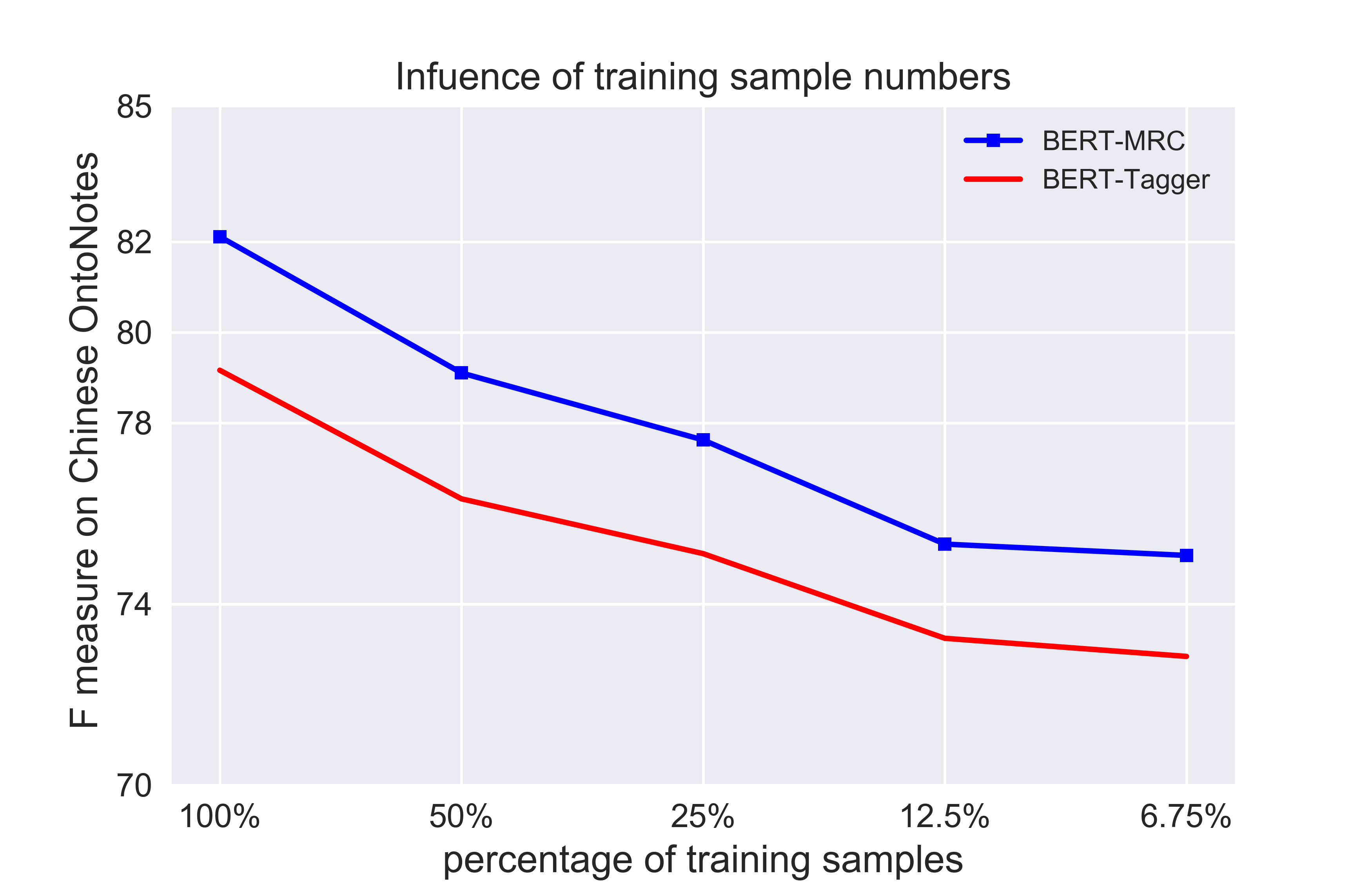}
\centering
\caption{
Effect of varying percentage of training samples on Chinese OntoNotes 4.0. BERT-MRC can achieve the same F1-score comparing to BERT-Tagger with fewer training samples. }
\label{train-samples}
\end{figure}

Since the natural language query encodes significant prior knowledge,  we expect that the proposed framework works better with less training data. 
Figure \ref{train-samples} verifies this point: 
on the Chinese OntoNotes 4.0 training set, 
the query-based BERT-MRC approach achieves comparable performance to BERT-tagger even with half amount of training data.


\section{Conclusion}
In this paper, we reformalize the NER task as a MRC question answering task. This formalization comes with two key advantages: (1) being capable of addressing overlapping or nested entities; (2) the query encodes significant prior knowledge about the entity category to extract. The proposed method obtains SOTA results on both nested and flat NER datasets, which indicates its effectiveness. 
In the future, we would like to explore variants of the model architecture. 

\section*{Acknowledgement}
We thank all anonymous reviewers, as well as Jiawei Wu and Wei Wu for their comments and suggestions. 
The work is supported by the National Natural Science Foundation of China (NSFC No. 61625107 and 61751209). 

\bibliography{ner_acl}
\bibliographystyle{acl_natbib}

\end{document}